\documentclass[conference]{IEEEtran}

\usepackage{mathtools}
\usepackage{pifont}
\usepackage{multirow}
\usepackage{enumitem}
\usepackage{booktabs}
\usepackage{subfigure}
\usepackage{caption}
\usepackage{gensymb}
\usepackage{cite}
\usepackage[labelfont=bf,font=small]{caption}
\def\BibTeX{{\rm B\kern-.05em{\sc i\kern-.025em b}\kern-.08em
    T\kern-.1667em\lower.7ex\hbox{E}\kern-.125emX}}

\usepackage[bookmarks=true,breaklinks=true,colorlinks,linkcolor=blue,citecolor=blue,urlcolor=blue]{hyperref}

\usepackage{comment}
\usepackage{xspace}
\newcommand{\carb}{IoTCO2}

\begin{document}
\title{\LARGE \bf \carb: End-to-End Carbon Footprint Assessment of Internet-of-Things-Enabled Deep Learning}

\author{
\begin{tabular}{cccc}
Fan Chen & Shahzeen Attari & Gayle Buck & Lei Jiang
\end{tabular}\\
Indiana University Bloomington\\
\{fc7,sattari,gabuck,jiang60\}@iu.edu
}

\maketitle

\begin{abstract}
To improve privacy and ensure quality-of-service (QoS), deep learning (DL) models are increasingly deployed on Internet of Things (IoT) devices for data processing, significantly increasing the carbon footprint associated with DL on IoT, covering both operational and embodied aspects. Existing operational energy predictors often overlook quantized DL models and emerging neural processing units (NPUs), while embodied carbon footprint modeling tools neglect non-computing hardware components common in IoT devices, creating a gap in accurate carbon footprint modeling tools for IoT-enabled DL. 
This paper introduces \textit{\carb}, an end-to-end tool for precise carbon footprint estimation in IoT-enabled DL, with deviations as low as 5\% for operational and 3.23\% for embodied carbon footprints compared to actual measurements across various DL models.
Additionally, practical applications of \carb~are showcased through multiple user case studies.

\end{abstract}


\IEEEpeerreviewmaketitle

\section{Introduction}

\begin{figure}[t!]
\centering
\begin{minipage}{0.5\linewidth}
\begin{center}
\includegraphics[width=1\linewidth]{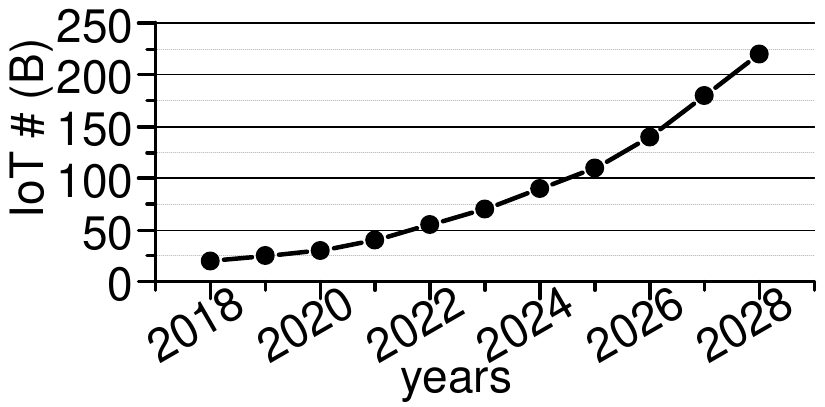}
\vspace{-0.2in}
\caption{Total IoT count.}
\label{f:carbon_iot_num}
\end{center}
\end{minipage}
\hspace{-8pt}
\begin{minipage}{0.5\linewidth}
\begin{center}
\includegraphics[width=1\linewidth]{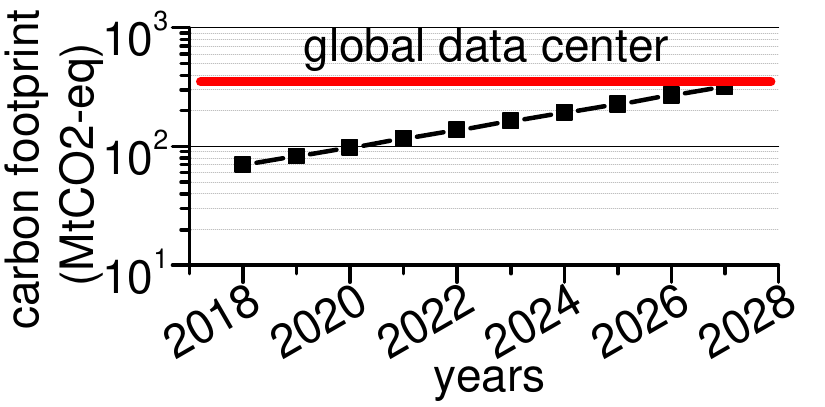}
\vspace{-0.2in}
\caption{IoT carbon footprint.}
\label{f:carbon_iot_carbon}
\end{center}
\end{minipage}
\vspace{-0.3in}
\end{figure}

Internet of Things (IoT)-enabled deep learning (DL)~\cite{Ignatov:ICCVW2019} has rapidly evolved, serving not only as a complement to cloud-based Machine Learning as a Service (MLaaS) but also as a superior platform for deploying DL models. It provides enhanced quality-of-service (QoS) and improved security for time-sensitive and privacy-critical applications
This shift is driving an exponential increase in the global number of IoT devices, projected to grow by approximately 40\% annually~\cite{sparks2017route}, as shown in Figure~\ref{f:carbon_iot_num}. 
However, this rapid growth comes with a substantial environmental cost, as the carbon footprint associated with IoT-enabled DL, both from \textit{operational} emissions due to daily usage and \textit{embodied} emissions from hardware manufacturing, is expected to rise dramatically. 
Alarmingly, as shown in Figure~\ref{f:carbon_iot_carbon}, it is projected that by 2028, carbon emissions from IoT devices could exceed those of global data centers~\cite{sparks2017route}, underscoring the urgent need for sustainable practices in this rapidly expanding field.

Unlike cloud-based MLaaS platforms that focus on DL training and typically rely on CPUs and GPUs, IoT devices primarily handle DL inference and utilize ARM-based System-on-Chip (SoC)~\cite{Ignatov:ICCVW2019}.
Moreover, while DL training commonly uses floating-point arithmetic, IoT-enabled DL predominantly employs Neural Processing Units (NPUs)~\cite{dally2020domain, Qualcomm:Gen3, Samsung:Exynos2100, Wegner:Pixel2023} within ARM-based SoCs, processing quantized DL inferences using INT16/8/4 arithmetic. However, existing DL carbon footprint estimation tools~\cite{Lacoste:ARXIV2019,Faiz:ICLR2024,Garcia:JPDC2019, Tu:SEC2023, Gupta:ISCA2022} fail to account for the specific characteristics of IoT devices, leading to limitations in accurately assessing both operational and embodied carbon emissions.

First,
previous operational carbon footprint estimators~\cite{Lacoste:ARXIV2019,Faiz:ICLR2024} have primarily focused on DL training, which relies on power-hungry GPUs and measures floating-point operations per second (FLOPs) required by various DL models. 
However, IoT devices primarily perform DL inference tasks using a variety of low-power hardware with limited concurrency. Relying solely on FLOPs results in suboptimal and inaccurate carbon footprint assessments. 
While some existing tools~\cite{Garcia:JPDC2019,Tu:SEC2023} can estimate the operational energy consumption of DL models on mobile CPUs and GPUs, they do not account for quantized ML models or NPUs~\cite{dally2020domain, Qualcomm:Gen3, Samsung:Exynos2100, Wegner:Pixel2023}, leading to significant deviations in operational carbon footprint for IoT devices.

Second,
previous efforts in embodied carbon footprint prediction~\cite{Gupta:ISCA2022} for DL have primarily focused on \textit{computing} components such as CPUs, GPUs, NPUs, DRAMs, and NAND flash memory. However, IoT devices also include various \textit{non-computing} hardware components~\cite{Pirson:JCP2021}, such as actuators, casings, printed circuit boards (PCBs), power supplies, analog sensors, and user interfaces (UIs). These non-computing components account for a significant portion, ranging from 30\% to 60\%, of the embodied energy in many IoT devices~\cite{Pirson:JCP2021}. 

In this paper, we introduce \carb, an end-to-end carbon footprint modeling framework for IoT devices executing DL inferences, with the following key contributions:

\begin{itemize}[leftmargin=*, nosep, topsep=0pt, partopsep=0pt]
\item \textbf{Operational Carbon Footprint}. 
We propose an operational carbon footprint estimator for IoT-enabled DL models in three steps: benchmarking inference energy across CPUs, GPUs, and NPUs; predicting energy via a random forest model; and linking predictions to carbon emissions. Unlike FLOPs-based methods for training, our approach focuses on inference, accounting for hardware diversity and NPU quantization, resulting in improved assessment accuracy.

\item \textbf{Embodied Carbon Footprint}. 
We present an accurate estimator for the embodied carbon footprint that models both computing and non-computing hardware components in IoT devices, ranging from low-power CPU controllers to high-end GPU platforms, ensuring comprehensive coverage.

\item \textbf{Accuracy \& User Case Studies}. 
\carb~demonstrates a minimum deviation of 5\% for operational and 3.23\% for embodied carbon footprints compared to actual measurements across various DL models on IoT devices.
We further illustrate the practical utility of \carb~through the examination of multiple user case scenarios, showcasing its applicability in real-world contexts.
\end{itemize}

\section{Background and Motivation}
\vspace{-0.04in}

\subsection{Background for IoT-Enabled Deep Learning}
\vspace{-0.04in}

\textbf{NPUs for Quantized DL Models}. 
In processing IoT-enabled DL~\cite{Ignatov:ICCVW2019}, ARM-based System-on-Chips (SoCs) are frequently preferred over traditional CPU or GPU cores due to several key factors. 
\underline{First}, 
both high-performance (i.e., large) and power-efficient (i.e., small) CPU cores are generally inefficient for DL inferences~\cite{dally2020domain}.
\underline{Second}, 
although mobile GPUs offer greater energy efficiency for certain inference tasks~\cite{Ignatov:ICCVW2019}, many low-end ARM-based SoCs lack the capability to support them. 
\underline{Third}, 
and more critically, most mobile GPUs are limited to floating-point arithmetic, making them unsuitable for the quantized DL models commonly used in IoT applications.
To address these challenges, there has been a growing trend towards incorporating Neural Processing Units (NPUs) within ARM-based SoCs. 
NPUs such as Qualcomm Hexagon~\cite{Qualcomm:Gen3}, Samsung Exynos~\cite{Samsung:Exynos2100}, and Google EdgeTPU~\cite{Wegner:Pixel2023} have been specifically designed to accelerate quantized DL inferences using INT16/8/4 arithmetic, significantly enhancing the efficiency of IoT-enabled DL applications.

\textbf{Operational Carbon Footprint Modeling}. 
Existing work~\cite{Lacoste:ARXIV2019,Faiz:ICLR2024,Garcia:JPDC2019, Tu:SEC2023} has significant limitations in modeling the operational carbon footprint for IoT-enabled DL.
\underline{First}, 
most tools~\cite{Lacoste:ARXIV2019,Faiz:ICLR2024} focus on DL \textit{training}, neglecting the more complex and frequent \textit{inference} phase. 
Unlike DL \textit{training}, which uses the Neural Scaling Law~\cite{Kaplan:ARXIV2020} to estimate floating-point operations per second (FLOPs) and calculate energy consumption on GPUs, \textit{inference} involves various low-power hardware such as mobile CPUs, GPUs, and NPUs. Relying solely on FLOPs leads to inaccurate estimations of the operational carbon footprint for IoT devices, as inference on these devices lacks parallelism and cannot be accurately modeled using traditional assumptions of concurrent processing.
\underline{Second}, 
while some tools~\cite{Garcia:JPDC2019,Tu:SEC2023} can predict the energy consumption of DL inference on mobile devices, they often overlook model quantization used in emerging NPUs~\cite{Ignatov:ICCVW2019, Qualcomm:Gen3, Samsung:Exynos2100, Wegner:Pixel2023}, resulting in significant discrepancies in energy estimation.
To address these challenges, \textit{accurately modeling the operational carbon footprint for IoT-enabled DL inference requires a methodology that accounts for diverse hardware, particularly NPUs running quantized models, and moves beyond traditional energy modeling based on neural scaling laws}.

\textbf{Embodied Carbon Footprint Modeling}. 
Previous efforts~\cite{Gupta:ISCA2022} to model the embodied carbon footprint have mainly focused on computing components such as CPUs, GPUs, NPUs, DRAMs, and NAND flash memory. 
These studies, however, tend to overlook the substantial impact of non-computing hardware components in IoT devices~\cite{Pirson:JCP2021}—such as actuators, casings, printed circuit boards (PCBs), power supplies, analog sensors, and user interfaces (UIs)—which often play a dominant role in the overall embodied carbon footprint of IoT devices~\cite{Sphera:GABI2023}.
\textit{Accurately modeling the embodied carbon footprint for IoT-enabled DL inference requires a comprehensive approach that incorporates both computing and non-computing components within an integrated framework}.

\subsection{Related Work and Motivation}
\vspace{-0.04in}
We summarize related work~\cite{Lacoste:ARXIV2019,Faiz:ICLR2024,Garcia:JPDC2019, Tu:SEC2023, Gupta:ISCA2022} and the motivated objectives of \carb~in Table~\ref{t:ml_related_compar}.
Unlike prior DL carbon footprint modeling tools~\cite{Lacoste:ARXIV2019,Faiz:ICLR2024} that primarily focus on the training phase, \carb~is motivated to shift the focus to the inference phase for various DL models deployed on resource-constrained IoT devices. 
While previous DL operational carbon footprint predictors~\cite{Garcia:JPDC2019,Tu:SEC2023} often overlook factors such as model quantization and NPU accelerators, 
\carb~is driven to incorporate these elements to provide more accurate estimates tailored to IoT-enabled DL.
Moreover, existing tools for embodied carbon footprint modeling~\cite{Gupta:ISCA2022} typically neglect non-computing components like actuators, casings, PCBs, power supplies, analog sensors, and UIs~\cite{Pirson:JCP2021}, leading to inaccuracies when applied to IoT devices. In response, \carb~is motivated to account for these components, aiming for more precise carbon footprint estimations for IoT-enabled DL.

\begin{table}[t!]
\caption{Comparison of prior works with~\carb.}
\label{t:ml_related_compar}
\vspace{-0.15in}

\setlength{\tabcolsep}{5.2pt}
\begin{center}

\begin{tabular}{|c||c|c|c|c|}\hline
\multirow{2}{*}{\textbf{Tool}} 
&\multirow{2}{*}{\textbf{Inference}}   
&\multicolumn{2}{|c|}{\textbf{Operational}} & \textbf{Embodied}\\\cline{3-5}
&  &\textbf{NPU} & \textbf{Quantization}  & \textbf{Non-Computing Parts}\\\hline\hline

\cite{Lacoste:ARXIV2019,Faiz:ICLR2024}   
&\ding{55}  &\ding{55}  &\ding{51}  &\ding{55} \\\hline

\cite{Garcia:JPDC2019,Tu:SEC2023}          
&\ding{51}  &\ding{55}  &\ding{55}  &\ding{55} \\\hline

\cite{Gupta:ISCA2022}
&\ding{55}  &\ding{55}  &\ding{55}  &\ding{55} \\\hline\hline

\textbf{\carb} 
&\ding{51}  &\ding{51}  &\ding{51}  &\ding{51} \\\hline
\end{tabular}

\end{center}
\vspace{-0.3in}
\end{table}

\begin{figure*}[t!]
\centering
\subfigure[Input Height and Width.]{
   \includegraphics[width=1.65in]{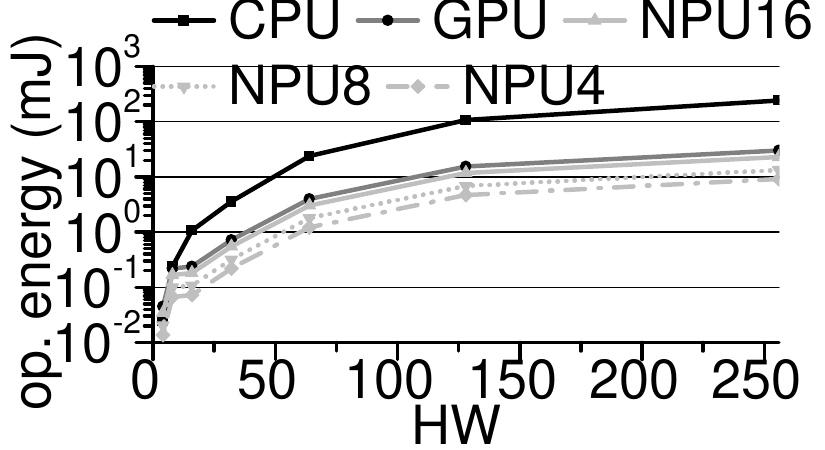}
   \label{f:iot_carbon_hw}
}
\hspace{-0.1in}
\subfigure[Kernel Size.]{
   \includegraphics[width=1.65in]{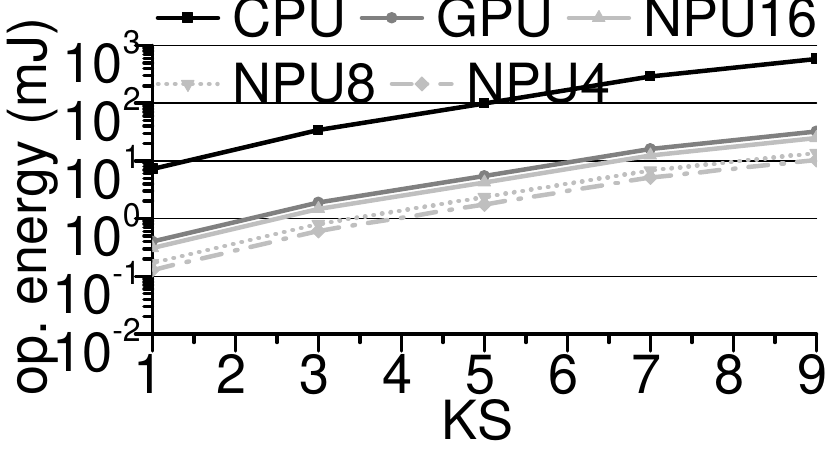}
   \label{f:iot_carbon_ks}
}
\hspace{-0.1in}
\subfigure[Input and Output Channel Number.]{
   \includegraphics[width=1.65in]{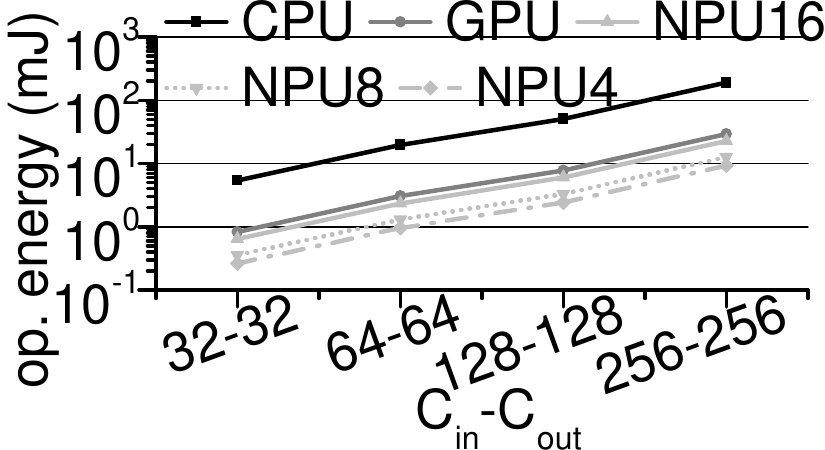}
   \label{f:iot_carbon_ch}
}
\hspace{-0.1in}
\subfigure[Stride Size.]{
   \includegraphics[width=1.65in]{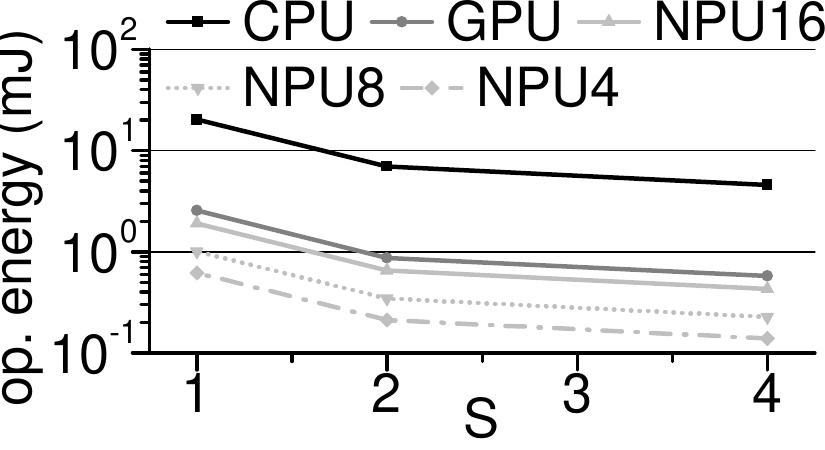}
   \label{f:iot_carbon_s}
}
\vspace{-0.1in}
\caption{The energy consumption of \texttt{conv+bn+relu} with various configurations on Snapdragon 8 G3.}
\label{f:iot_carbon_measure}
\vspace{-0.27in}
\end{figure*}

\section{\carb}
In this section, we present the \carb~framework, specifically developed to address and overcome the limitations in end-to-end modeling of both operational and embodied carbon footprints in IoT-enabled deep learning inference.

\vspace{-0.05in}
\subsection{Modeling Operational Carbon Footprint}
\vspace{-0.03in}

\textbf{Overview}.
\carb~introduces an operational carbon footprint model for DL inference in IoT devices. Since FLOPS, commonly used for DL training, is not suitable for inference, we use a more accurate approach with DL inference kernels. 
The process involves three steps:
(1) benchmarking energy consumption for DL inference kernels across mainstream IoT platforms, including CPUs, GPUs, and NPUs; 
(2) applying a random forest regression model trained on the obtained limited benchmarks to predict DL energy consumption; 
and (3) linking the predicted energy consumption to carbon emissions.

\subsubsection{Benchmarking DL Inference Kernels}
Mobile DL compilers~\cite{Ignatov:ICCVW2019} optimize inference kernels, the foundational units in DL inference, unlike in the training phase. The types and configurations of these kernels significantly impact operational energy consumption. A DL model comprises various kernels, such as convolution (\texttt{conv}), depthwise convolution (\texttt{dwconv}), activations (\texttt{relu}), pooling (\texttt{avg/maxpool}), batch normalization (\texttt{bn}), and fully connected layers (\texttt{fc}). To enhance energy efficiency, multiple kernels are often fused into single operations; for instance, \texttt{conv}, \texttt{bn}, and \texttt{relu} can be combined into one operation (\texttt{conv+bn+relu}), reducing memory accesses and improving performance. The following provides details on the measurement setup and kernel-level energy benchmarking for these DL inference kernels.

\begin{itemize}[leftmargin=*, nosep, topsep=0pt, partopsep=0pt]

\item \textbf{Measurement Setup}. 
We selected two leading IoT SoCs, the Qualcomm Snapdragon 8 G3~\cite{Qualcomm:Gen3} and Samsung Exynos 2100~\cite{Samsung:Exynos2100}, for evaluation, with configurations detailed in Table~\ref{t:ml_oper_config}. 
We isolate the battery management system chip from the SoC, utilizing it as a bridge to connect the SoC to an external power monitor. 
To minimize extraneous power consumption, we optimize device settings by reducing screen brightness, lowering the refresh rate, enabling flight mode, and disabling background applications. Power measurements were conducted with room temperatures between $20\degree$ to $25\degree$, using a power monitor with a 5 kHz sampling rate. Inferences were executed using the Qualcomm Neural Processing SDK and the Samsung ONE library.

\item \textbf{Kernel-Level Energy Measurement}. 
Table~\ref{t:ml_kernel_profile} summarizes the fused inference kernels on two candidate IoT SoCs, detailing parameters such as input/output dimensions ($HW$), channel numbers ($C_{i/o}$), kernel size ($KS$), stride size ($S$), and bitwidth ($BW$). As a representative example, we present the measured results for the NPUs. Notably, while NPUs are optimized for INT4, INT8, and INT16, mobile CPUs and GPUs uniformly use a 32-bit width, optimized for floating-point operations.
We employed adaptive sampling~\cite{Zhang:SIGCOMM2021} to randomly select advantageous kernel configurations from the full configuration space. Subsequently, we measured the average power consumption and inference latency for each kernel on the respective computing units, with each measurement averaged over 500 runs, using a single large CPU core (e.g., X4 or X1).

\end{itemize}

\vspace{4pt}
\subsubsection{DL Inference Energy Prediction}
The size of the complete kernel sampling space is calculated by multiplying all configuration dimensions. For instance, the sample space for the \texttt{conv+bn+relu} kernel encompasses all possible configurations of $HW$, $C_i$, $C_o$, $KS$, $S$, and $BW$, resulting in millions of combinations, which far exceed practical sampling capabilities.
To address this, we employ Random Forest Regression (RFR)~\cite{Breiman:ML2001}, an ensemble decision tree-based learning algorithm. Our kernel-level energy predictor utilizes individual RFR models for each kernel type, trained with collected kernel-level energy data shown in Table~\ref{t:ml_kernel_profile}. 
The total energy consumption of DL inference is estimated by aggregating the energy consumption of all kernels within the neural network. The RFR energy predictor accepts the neural network architecture, characterized by various kernel configurations, as input and produces the predicted operational energy of the inference as output.

\textbf{Kernel Energy Analysis}. 
Among all computing units, we find that the \texttt{conv+bn+relu} kernel exhibits the highest operational energy consumption. Figure~\ref{f:iot_carbon_measure} illustrates the impact of varying configurations of the \texttt{conv+bn+relu} kernel on its energy consumption in the Snapdragon 8 G3, where one parameter is varied while others remain constant. The relationship between energy consumption and kernel configurations is non-linear, particularly for this kernel type, and cannot be characterized by a simple equation. Increasing the value of a configuration parameter generally leads to higher energy consumption, with \texttt{KS} being especially sensitive due to its quadratic relationship with convolution time complexity. Throughout parameter exploration, the mobile CPU consistently exhibits the highest energy consumption, while the mobile GPU reduces operational energy consumption by $1.1\times$ to $18\times$. By quantizing the same kernels to INT16, INT8, and INT4, the mobile NPU achieves further reductions in energy consumption compared to the mobile GPU, averaging decreases of $23\%$, $58\%$, and $70\%$, respectively.

\begin{table}[t!]
\vspace{0.04in}
\caption{Specifications of the two representative target IoT SoCs.}
\vspace{-0.15in}
\setlength{\tabcolsep}{4.8pt}
\label{t:ml_oper_config}
\begin{center}
\begin{tabular}{|c||c|c|c|}\hline
\textbf{SoC}  
& \textbf{ARM Cortex CPU}  & \textbf{GPU}  & \textbf{NPU}  \\\hline\hline

Snapdragon 8 G3  
& 1 X4, 5 A720, 2 A520  & Adreno 750 & Hexagon   \\\hline

Exynos 2100  
& 1 X1, 3 A78, 4 A55      & Mali G78      & 3-core NPU  \\\hline

\end{tabular}
\end{center}
\vspace{-0.22in}
\end{table}
\begin{table}[t!]
\vspace{0.05in}
\caption{The kernels measured on NPUs.}
\vspace{-0.15in}
\setlength{\tabcolsep}{1pt}
\label{t:ml_kernel_profile}
\begin{center}
\begin{tabular}{|c||c|c|c|}\hline
\textbf{Kernel}
&\textbf{Num.}     
&\textbf{Engy.} ($mJ$)               
&\textbf{Configurations}
\\\hline\hline

\texttt{conv+bn+relu}    &3096  &0.001$\sim$108.2  &($HW$, $C_{i}$, $C_{o}$, $KS$, $S$, $BW$) \\\hline
\texttt{dwconv+bn+relu}  &1047  &0.008$\sim$0.61   & ($HW$, $C_{i}$, $KS$, $S$, $BW$) \\\hline
\texttt{bn+relu}         &300   &0.001$\sim$12.21  & ($HW$, $C_{i}$, $BW$) \\\hline
\texttt{relu}           &138    &0.002$\sim$6.32   & ($HW$, $C_{i}$, $BW$) \\\hline
\texttt{avg/max pool}   &84     &0.014$\sim$1.312  & ($HW$, $C_{i}$, $KS$, $S$, $BW$) \\\hline
\texttt{fc}            &72      &0.002$\sim$32.23  & ($C_{i}$, $C_{o}$, $BW$) \\\hline
\texttt{concat}         &426    &0.043$\sim$3.12   & ($HW$, $C_{i1}$, $C_{i2}$, $C_{i3}$, $C_{i4}$, $BW$) \\\hline
\texttt{others}        &294     &0.001$\sim$12.38  & ($HW$, $C_{i}$, $BW$) \\
\hline
\end{tabular}
\end{center}
\vspace{-0.36in}
\end{table}

\vspace{4pt}
\subsubsection{Linking Operational Energy to Carbon}
To estimate the operational carbon footprint of IoT-enabled deep learning, 
we utilize Carbon Intensity (CI), a widely recognized metric that quantifies the environmental impact of electricity based on its cleanliness. Each electric grid has a distinct CI value, determined by factors such as transmission efficiency and the composition of energy sources, including hydro, wind, coal, or gas. For instance, the CI values~\cite{Report:28} for Australia, the UK, and France are 0.656, 0.281, and 0.054 $gCO2$-$eq/kWh$, respectively. Assuming IoT devices draw power from local electric grids, the operational carbon footprint of IoT-enabled DL ($CO2eq_{op}$) can be calculated as shown in Equation~\ref{e:co2_op}, where $energy_{op}$ represents the operational energy of a DL inference, $CI$ is the carbon intensity for the IoT device's location, and $num$ indicates the number of DL inferences.
It is important to note that by considering the specific energy sources that power the IoT devices, the relationship between operational energy consumption and carbon emissions can be further refined, enabling a more accurate calculation.

\vspace{-12pt}
\begin{equation}
CO2eq_{op} = energy_{op} \times CI \times num
\label{e:co2_op}
\end{equation}

\subsection{Modeling Embodied Carbon Footprint}
\textbf{Overview}.
\carb~quantifies the embodied carbon footprint ($CO2eq_{emb}$) of an IoT device, encompassing both computing and non-computing components, using equation~\ref{e:carbon_all_device}. 
In this equation, $CO2eq_{component_i}$ represents the embodied carbon footprint of hardware component $i$, categorized as either computing ($cmp$) or non-computing ($ncmp$). The term $time_i$ indicates the active usage duration of component $i$, while $lifetime$ refers to the expected lifespan of the IoT device. For instance, Google assumes a 3-year lifespan for all its IoT devices~\cite{Google:SUS2024}. 
In the following, we outline the methods for evaluating the carbon footprints of computing and non-computing components in IoT devices, respectively.
\vspace{-4pt}
\begin{equation}
\displaystyle CO2eq_{emb} = \sum_{i\in cmp || ncmp} \frac{time_i\cdot CO2eq_{component_i}}{lifetime}
\label{e:carbon_all_device}
\end{equation}
\vspace{-8pt}

\subsubsection{Evaluation of Computing Components}
The embodied carbon footprint of computing components primarily arises from the logic computing chip ($CO2eq_{log}$) and the memory chip ($CO2eq_{mem}$)~\cite{Gupta:ISCA2022}.
$CO2eq_{log}$ can be calculated using Equation~\ref{e:carbon_all_chip}, where $CPA$ denotes the carbon emissions per unit area, and $area$ represents the chip's total area.
$CO2eq_{mem}$ encompassing both DRAM and Flash memory, can be determined using Equation~\ref{e:carbon_all_mem}, where $CPC$ signifies the carbon emissions per gigabyte, and $capacity$ represents the memory capacity. 
The values of $CPA$ and $CPC$ depend on various semiconductor fabrication factors, including yield, energy consumption per unit area during manufacturing, emissions from the use of chemicals in production, and emissions associated with sourcing raw materials.
Trends in $CPA$, DRAM $CPC$, and Flash $CPC$ across different process technologies are illustrated in Figures~\ref{f:carbon_logic_cpa},~\ref{f:carbon_dram_cpa}, and~\ref{f:carbon_ssd_cpa}, respectively.

\vspace{-10pt}
\begin{equation}
CO2eq_{log} = CPA \cdot area
\label{e:carbon_all_chip}
\end{equation}
\vspace{-15pt}
\begin{equation}
CO2eq_{mem} = CPC \cdot capacity
\label{e:carbon_all_mem}
\end{equation}
\vspace{-15pt}

\subsubsection{Evaluation of Non-computing Components}
The embodied carbon footprint of non-computing components in IoT devices has been largely overlooked in previous research, despite being significantly more complex to assess than that of computing components. This complexity arises from the diverse range of non-computing elements, as outlined in Table~\ref{t:carb_non_comput}. 
In the following, we outline the methodology for calculating the embodied carbon footprint of each type of non-computing component. 
Most carbon-related data for these components are sourced from the Sphera GaBi Extension Database XI - Electronics~\cite{Sphera:GABI2023} and relevant industry reports~\cite{Google:Watch2022, Google:TV2022,Google:Phone2021}.

\begin{itemize}[leftmargin=*, nosep, topsep=0pt, partopsep=0pt]

\item \textbf{Actuator}.  
We consider three types of actuators: vibration motors, solid-state relays (SSRs), and DC motors. The manufacturing of each vibration motor/SSR/DC motor incurs carbon emissions of 0.03/0.165/1.03 kgCO2-eq. An IoT device typically incorporates multiple actuators.

\item \textbf{Casing}. 
The casing of an IoT device comprises ABS plastic granules, aluminum, steel, and glass. The embodied carbon of these materials is calculated by Equation~\ref{e:carbon_all_non}, where $CPM$ is the carbon emissions per unit mass, and $mass$ represents the material mass. The $CPM$ values for ABS plastic granules, aluminum, steel, and glass are 0.08, 0.02, 0.06, and 0.02 kgCO2-eq/g, respectively.

\vspace{-10pt}
\begin{equation}
CO2eq_{non} = CPM \cdot mass,
\label{e:carbon_all_non}
\end{equation}
\vspace{-20pt}
\begin{equation}
CO2eq_{t} = CPD \cdot mass \cdot distance
\label{e:carbon_all_tran}
\end{equation}
\vspace{-15pt}

\begin{figure}[t!]
\centering
\subfigure[Logic.]{
   \includegraphics[width=1.2in]{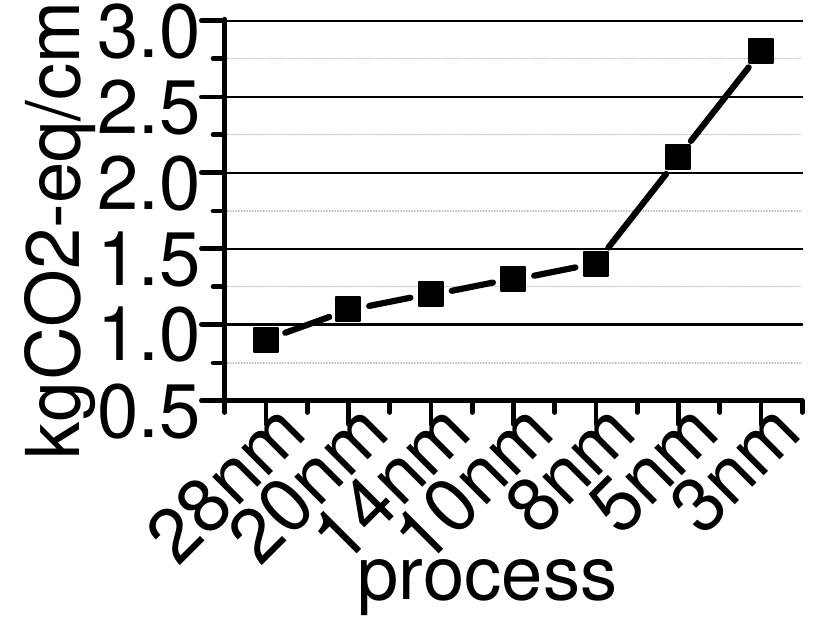}
   \label{f:carbon_logic_cpa}
}
\hspace{-0.15in}
\subfigure[DRAM.]{
   \includegraphics[width=1.05in]{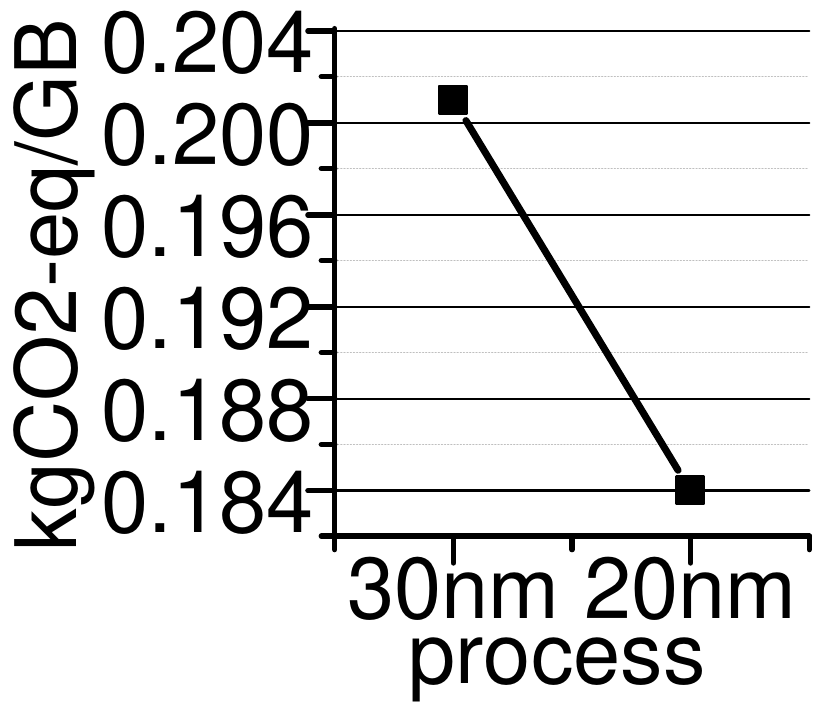}
   \label{f:carbon_dram_cpa}
}
\hspace{-0.15in}
\subfigure[Flash.]{
   \includegraphics[width=1.05in]{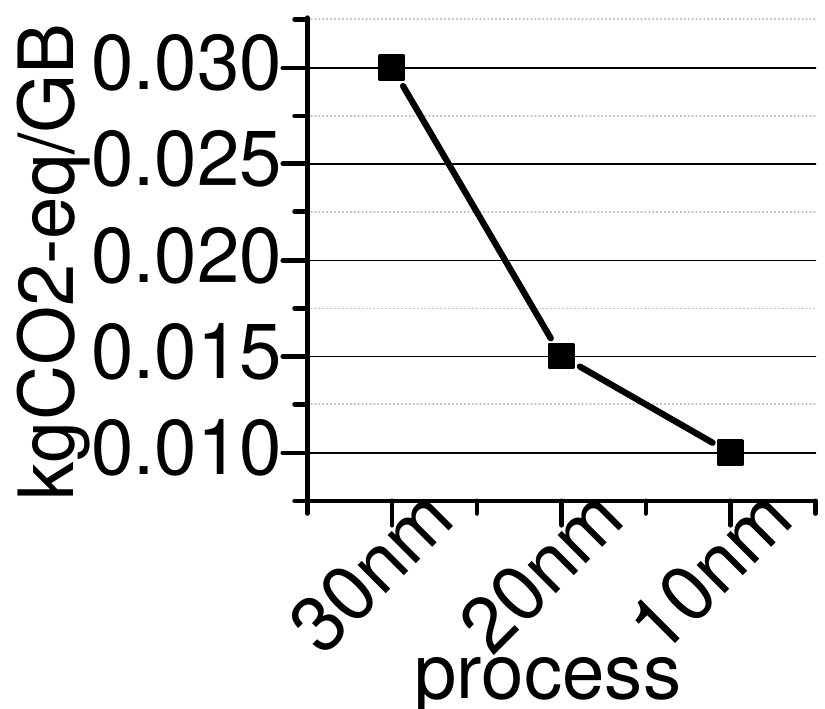}
   \label{f:carbon_ssd_cpa}
}
\vspace{-0.1in}
\caption{The embodied carbon footprint of chip fabrication.}
\label{f:carb_chip_fab}
\vspace{-0.12in}
\end{figure}
\begin{table}[t!]
\caption{The description of non-computing components.}
\vspace{-0.08in}
\setlength{\tabcolsep}{4pt}
\label{t:carb_non_comput}
\begin{center}
\begin{tabular}{|l||l|}\hline
\textbf{Component} &\textbf{Description} \\\hline\hline
Actuator  &Vibration Motor; SSR; DC Motor \\\hline
Casing    &ABS Granules; Aluminum; Steel; Glass  \\\hline
Connectivity      &Whip-Like Antenna \\\hline
PCB   &FR4; Solder Paste \\\hline
Power Suply    &Coin Cell; AA/A Alkaline; Li-Ion Battery \\\hline
Sensing   &Analog Sensors; Microphone \\\hline
UI       &LED; LCD Screen \\\hline
Transportation &Truck; Plane \\\hline
Others    & Capacitor; Resistance; Diode; Transistor; Cables \\\hline
\end{tabular}
\end{center}
\vspace{-0.3in}
\end{table}

\item \textbf{Connectivity}.  
The connectivity of an IoT device can be facilitated by a whip-like antenna. The embodied carbon emissions of an antenna are computed using Equation~\ref{e:carbon_all_non}, with a $CPM$ value of 0.019kgCO2-eq/g.

\item \textbf{PCB}.  
FR4, a dielectric material, is used in PCB fabrication.
For 4- or 8-layer PCBs, embodied carbon emissions are calculated using Equation~\ref{e:carbon_all_chip}. The $CPA$ value for an 8-layer PCB with solder pastes is 0.032kgCO2-eq/c$m^2$.

\item \textbf{Power Supply}. 
IoT power supplies include coin cell Li-Po, AA/AAA alkaline, and Li-ion batteries, with embodied emissions of 0.02, 0.38, and 0.18 kgCO2-eq, respectively.
The embodied carbon of a Li-ion battery, calculated using Equation~\ref{e:carbon_all_non}, has a $CPM$ of 0.25 kgCO2-eq/g.

\item \textbf{Sensing}. 
IoT devices commonly use electret microphones and analog sensors for sensing. The embodied carbon emissions of an electret microphone, calculated using Equation~\ref{e:carbon_all_non}, have a $CPM$ of 0.2 kgCO2-eq/g. For analog sensors, emissions are computed using Equation~\ref{e:carbon_all_chip}, with $CPA$ values ranging from 0.8 to 2.46kgCO2-eq/c$m^2$.

\item \textbf{UI}. 
LEDs and LCD screens are often employed as user interfaces (UI) in IoT devices. The embodied carbon of LEDs is 0.03 kgCO2~\cite{Sphera:GABI2023}. For LCD screens, the embodied carbon emissions are calculated using Equation~\ref{e:carbon_all_non}, with a $CPM$ value of 0.11 kgCO2-eq/g.

\item \textbf{Transportation}. 
Embodied carbon emissions from transporting IoT devices are calculated using Equation~\ref{e:carbon_all_tran}, with $CPD$ values varying by mode: 8.68E-09 kgCO2-eq/g/km for trucks and 5.90E-07 kgCO2-eq/g/km for planes.

\item \textbf{Others}. 
Other non-computing components may include capacitors, resistors, diodes, transistors, crystals, and cables. The embodied carbon emissions for these materials are derived using Equation~\ref{e:carbon_all_non}, assuming an average $CPM$ value of 0.06 kgCO2-eq/g.
\end{itemize}

\begin{table*}[t!]
\caption{Neural networks inferring ImageNet (meas.: measured; est. $\Delta$: estimation deviation).}
\vspace{-0.15in}
\setlength{\tabcolsep}{4.8pt}
\label{t:carb_net_arch}
\begin{center}
\begin{tabular}{|c|c|c|c|c|c||c|c|c|c|c|c|c|c|c|c|}\hline

\multicolumn{6}{|c||}{\textbf{Validation Setup}}
& \multicolumn{10}{c|}{\textbf{Validation Results}} \\\hline

\multirow{2}{*}{\textbf{Network}}   
&\multirow{2}{*}{\textbf{FLOPs}} 
&\multicolumn{4}{c||}{\textbf{Top-1 Accuracy} (\%)}  
&\multicolumn{2}{c|}{\textbf{CPU} ($mJ$)}  
&\multicolumn{2}{c|}{\textbf{GPU} ($mJ$)}  
&\multicolumn{2}{c|}{\textbf{NPU16} ($mJ$)}  
&\multicolumn{2}{c|}{\textbf{NPU8} ($mJ$)}  
&\multicolumn{2}{c|}{\textbf{NPU4} ($mJ$)}
\\\cline{3-16}
& &FP32 &INT16 &INT8 &INT4  
&meas. &est. $\Delta$ 
&meas. &est. $\Delta$ 
&meas. &est. $\Delta$ 
&meas. &est. $\Delta$  
&meas. &est. $\Delta$
\\\hline\hline

MobileNetv2  &585M &71.7 &71.2 &70.8 &68.2  
&0.172 &$\mathbf{+10\%}$ &0.113 &$\mathbf{+16\%}$ &0.104 &$\mathbf{+9\%}$ &0.082 &$\mathbf{+13\%}$ &0.071 &$\mathbf{+21\%}$ \\\hline

ShuffleNetv2 &149.7M &68.6 &67.9 &67.5 &66.8 
&0.075 &$\mathbf{-16\%}$ &0.057 &$\mathbf{-19\%}$ &0.054 &$\mathbf{-14\%}$ &0.043 &$\mathbf{-6\%}$ &0.037 &$\mathbf{-19\%}$ \\\hline

SqueezeNet   &352M	 &59.9 &59.6 &59.1 &57.5      
& 0.139  &$\mathbf{-13\%}$ &0.116 &$\mathbf{-11\%}$ &0.112 &$\mathbf{-16\%}$ &0.089 &$\mathbf{-9\%}$ &0.078 &$\mathbf{-15\%}$ \\\hline

ResNet18     &1.8G   &71.5 &70.2 &69.5 &68.8  
&0.621 &$\mathbf{+8\%}$ &0.333 &$\mathbf{+5\%}$ &0.301 &$\mathbf{+6\%}$ &0.233 &$\mathbf{+12\%}$ &0.201 &$\mathbf{+18\%}$ \\\hline

\end{tabular}
\end{center}
\vspace{-0.25in}
\end{table*}

\vspace{-0.06in}
\section{\carb~Validation}
\vspace{-0.02in}
In this section, we validate the proposed \carb~framework by presenting comprehensive results for both operational and embodied carbon footprints. We outline the validation setup, detailing the selection of representative DL models, IoT devices, and the methodologies used to obtain measurements. \carb's estimates are then compared against ground truth data, with a thorough analysis and discussion of the results.

\vspace{-0.06in}
\subsection{Validation of Operational Carbon Footprint}
\vspace{-0.02in}

\textbf{Validation Setup}. 
To validate the operational carbon emission modeling, we selected four neural networks as detailed in the left section of Table~\ref{t:carb_net_arch}. 
These networks—MobileNetv2, ShuffleNetv2, SqueezeNet, and ResNet18—were all trained on the ImageNet dataset and consist of 3.5M, 2.3M, 1.23M, and 11.7M weights, respectively. 
Each network processes a $224\times224$ image, utilizing different numbers of FLOPs and achieving distinct top-1 accuracies. While the floating-point version (FP32) of each neural network provides the highest accuracy, model quantization only marginally reduces this accuracy. For example, even the most aggressive INT4 quantization results in only a 2.6\% to 4.2\% decrease in top-1 accuracy. Given this minimal accuracy loss, designing NPUs optimized for quantized inference can significantly reduce operational energy consumption while maintaining competitive accuracy. In this validation, we assume all network inferences are executed on the Snapdragon 8 G3, and we predict their corresponding operational energy consumption.

\textbf{Validation Results}.
The operational energy validation results are presented in the right part of Table~\ref{t:carb_net_arch}, where the measured (meas.) operational energy values for DL inferences across various network architectures serve as the ground truth reference. NPU$X$ denotes the execution of an INT$X$-quantized inference on the NPU. Unlike previous studies~\cite{Tu:SEC2023}, our predictor accurately models the operational energy of quantized inferences on an NPU. Overall, the predicted operational carbon footprint for DL inferences shows high accuracy, with estimation deviations (est. $\Delta$) ranging from a minimum of +5\% to a maximum of +21\% from the ground truth.

\vspace{-0.06in}
\subsection{Validation of Embodied Carbon Footprint}
\vspace{-0.02in}

\textbf{Validation Setup}. 
To validate the embodied carbon emission modeling of \carb, we have selected three representative IoT devices: the low-end Chromecast Google TV, the middle-end Google Pixel Watch 2, and the high-end Google Pixel 6 Pro Phone. The hardware configurations of these IoT devices are provided in Table~\ref{t:carb_embod_val2}. Comprehensive hardware and chip specifications for these IoT devices are extracted from sources such as~\cite{Aufranc:CNX2019},~\cite{Simic:Qualcomm2022}, and~\cite{Wegner:Pixel2023}. Specifically, the chipset of each IoT device includes multiple components beyond the primary CPU SoC chip. For instance, the chipset of the Google Pixel 6 Pro Phone comprises a range of components, including the 5G Modem, RF/SM transceiver, power management integrated circuit, NFC controller, surge protection, battery charger IC, security processor, audio amplifier, GNSS receiver IC, wireless charging receiver IC, Wi-Fi/BT module, supply modulator, WLAN front-end modules, and a phased array of microcoils~\cite{Wegner:Pixel2023}. The inclusion of these diverse components results in the extensive area and complexity associated with the chipset of the Google Pixel 6 Pro Phone.

\begin{table}[t!]
\caption{IoT device setup for carbon footprint validation.}
\vspace{-0.15in}
\setlength{\tabcolsep}{4.8pt}
\label{t:carb_embod_val2}
\begin{center}
\begin{tabular}{|c||l|}\hline
\textbf{Device}
& \textbf{Hardware Configuration}
\\\hline\hline

& Snapdragon W5+~\cite{Simic:Qualcomm2022}, 2GB DRAM, 32GB Flash,  \\
Google Pixel  & 4nm CPU Cortex-A53 and GPU A702, 22nm \\
Watch 2~\cite{Google:Watch2022}		& Cortex-M55,  PCB, 300$mm^2$, chipset 200$mm^2$,\\
(\textit{Mid-End}) &SoC 90$mm^2$, plastic 36g, steel 15g, aluminum\\
&4g, battery 5g, LCD 6g, other 1g \\\hline\hline
 
& Amlogic S905X3~\cite{Aufranc:CNX2019}, 2GB DRAM, 8GB Flash, \\
Chromecast HD &22nm CPU Cortex-A55 and Cortex-M3, GPU\\
Google TV~\cite{Google:TV2022}     & Mali G31MP2, 1.2 TOPS NN inference acc., \\
(\textit{Low-End}) &PCB 99$mm^2$, chipset 97$mm^2$, SoC 110 $mm^2$,\\
&plastic 56g, steel 4g, aluminum 11g, other 6g\\\hline\hline

&Google Tensor~\cite{Wegner:Pixel2023}; 12GB DRAM; 128GB Flash,  \\
Google Pixel 6  &5nm CPU Cortex-A78, and Cortex-A55,  GPU Mali\\
Pro Phone~\cite{Google:Phone2021}   & G78, DL acc., PCB 1200$mm^2$, chipset 714$mm^2$, \\
(\textit{High-End}) &SoC 110$mm^2$, plastic 17g, steel 15g, alum. 32g, \\
&battery 64g, LCD 25g, glass 17, other 21g\\\hline

\end{tabular}
\end{center}
\vspace{-0.1in}
\end{table}
\begin{table}[t!]
\caption{Validation of our $CO2eq_{emb}$ model (in kgCO2-eq.)}
\vspace{-0.15in}
\setlength{\tabcolsep}{4.8pt}
\label{t:carb_embod_val}
\begin{center}
\begin{tabular}{|c||c|c|c|}\hline
\multirow{2}{*}{\textbf{Component}}    
& \textbf{Google Pixel}  
& \textbf{Chromecast HD} 
& \textbf{Google Pixel 6}  \\
                                                
& \textbf{Watch 2}         
& \textbf{Google TV}         
& \textbf{Pro Phone}        
\\\hline\hline

SoC &2.7 &0.99 &3.67  \\\hline
Chipset &2.8 &0.5 &23.8  \\\hline
DRAM &1.2 &1.2 &7.2  \\\hline
Flash &0.96  &0.24 &3.84 \\\hline
PCB  &3.6  &1.19 &14.4 \\\hline
Casing &3.92 &5.48 &4.79 \\\hline
Battery &1.25 &0.8 &16 \\\hline
LCD &0.66  &0 &2.75 \\\hline
Other &0.06 &0.54 &1.89 \\\hline\hline
Total &17.1  &10.4 &76.8 \\\hline
Google Reported &17.43 &11.1 &77.9  \\\hline
Est. $\mathbf{\Delta}$  &\textbf{-1.95\%}  &\textbf{-6.32\%} &\textbf{-1.43\%}                    \\\hline
\end{tabular}
\end{center}
\vspace{-0.35in}
\end{table}

\textbf{Validation Results}. 
The validation outcomes are detailed in Table~\ref{t:carb_embod_val}, where the reported values, serving as the ground truth, represent the embodied carbon footprints of the selected IoT devices. These values were sourced from official Google sustainability reports~\cite{Google:TV2022,Google:Watch2022,Google:Phone2021}. On average, the embodied carbon emissions estimated by \carb~demonstrate a remarkable alignment with the reported values, with a deviation of only 3.23\% on the lower side. The maximum observed discrepancy is a difference of -6.32\% between the \carb~estimates and the reported values. This close correspondence underscores the robustness and accuracy of the \carb~model in estimating the embodied carbon emissions for various IoT devices.

\section{Case Studies Using \carb}

We employed \carb~to conduct the following case studies.

\begin{figure}[t!]
\centering
\includegraphics[width=3.3in]{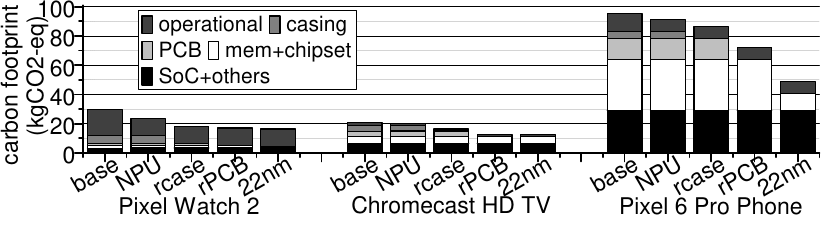}
\vspace{-0.05in}
\caption{The IoT device carbon footprint breakdown (\textbf{NPU}: employing NPU; rcase: recycled casing; rPCB: recycled PCB; and 22nm: fabricated by the 22nm process technology).}
\label{f:carbon_user_all}
\vspace{-0.2in}
\end{figure}

\textbf{Carbon Footprint Breakdown}. 
Using data on usage frequencies, transportation distances, and recycling overhead from Google sustainability reports~\cite{Google:TV2022,Google:Watch2022,Google:Phone2021}, we present the carbon footprint breakdown for the Google Pixel Watch 2, Chromecast Google TV, and Google Pixel 6 Pro Phone over a 3-year lifespan, as estimated by \carb~in Figure~\ref{f:carbon_user_all}, where "operational" represents the operational carbon footprint, while all other categories reflect the embodied carbon footprint. We assume all devices are used exclusively for DL inferences, with the UK's CI factor of 0.281 gCO2-eq/kWh applied.
The results indicate that for low-end IoT devices like the Chromecast Google TV, operational carbon emissions are the primary contributor to the overall carbon footprint. In contrast, for mid- and high-end devices such as the Google Pixel Watch 2 and Google Pixel 6 Pro Phone, the embodied carbon footprint plays a more significant role.

\textbf{Operational Carbon Footprint Mitigation}. 
As shown in Figure~\ref{f:carbon_user_all}, 
integrating NPUs into IoT SoCs does not notably increase the device's embodied carbon footprint, which is primarily dominated by components such as casings, PCBs, DRAMs, and peripheral chipsets. However, executing INT4-quantized inferences on NPUs can significantly reduce operational carbon emissions, with reductions of 20\% for the Chromecast Google TV and 4.3\% for the Google Pixel 6 Pro Phone. In contrast, the Google Pixel Watch 2 sees no appreciable decrease in carbon emissions with this approach.

\textbf{Embodied Carbon Footprint Reduction}.  
The embodied carbon footprints of the Google Pixel Watch 2, Chromecast Google TV, and Google Pixel 6 Pro Phone constitute 39.7\%, 91\%, and 87\% of their total emissions, respectively. For middle- and low-end IoT devices like the Google Pixel Watch 2 and Chromecast Google TV, aesthetically pleasing casings contribute significantly to their embodied carbon footprints. Utilizing fully-recycled casings (rcase) could reduce carbon emissions by 23\%, 19\%, and 5.2\%, respectively. Additionally, PCBs are another significant contributor to embodied carbon emissions. Employing fully-recycled PCBs (rPCB) could potentially reduce the carbon footprint of the Google Watch, TV, and phone by 6.5\%, 22\%, and 16.6\%, respectively. Furthermore, peripheral chipsets, DRAMs, and Flashes also contribute substantially to the embodied carbon footprint of these IoT devices. Transitioning from 5nm to 22nm technology for chipset, DRAM, and Flash fabrication could result in a 4.1\% and 32\% reduction in total carbon emission for the Google Pixel Watch 2 and Google Pixel 6 Pro Phone.


\section{Conclusion}
In this paper, we introduce \carb, an end-to-end modeling tool for assessing the carbon footprint of IoT-enabled DL applications. As IoT-enabled DL is projected to significantly increase global carbon emissions in the coming decade, \carb~offers accurate evaluations of both operational and embodied carbon footprints. By providing precise assessments, \carb~supports the implementation of effective measures to reduce carbon footprints. Across various DL models, \carb~demonstrates a maximum deviation of $\pm21\%$ compared to actual measurements.

\bibliographystyle{ieeetr}
\bibliography{co2}

\end{document}